# Causal Analysis and Classification of Traffic Crash Injury Severity Using Machine Learning Algorithms

Meghna Chakraborty[1], Timothy Gates, Subhrajit Sinha

*Department of Civil and Environmental Engineering, Michigan State University, 428 South Shaw Lane, East Lansing, MI 48824, USA, Email: chakra43@msu.edu*

*Department of Civil and Environmental Engineering, Michigan State University, 428 South Shaw Lane, East Lansing, MI 48824, USA, Email: gatestim@egr.msu.edu*

*Pacific Northwest National Laboratory, 902 Battelle Blvd, Richland, WA 99354, subhrajit.sinha@pnnl.gov*

**Abstract**

Causal analysis and classification of injury severity applying non-parametric methods for traffic crashes has received limited attention. This study presents a methodological framework for causal inference, using Granger causality analysis, and injury severity classification of traffic crashes, occurring on interstates, with different machine learning techniques including decision trees (DT), random forest (RF), extreme gradient boosting (XGBoost), and deep neural network (DNN). The data used in this study were obtained for traffic crashes on all interstates across the state of Texas from a period of six years between 2014 and 2019. The output of the proposed severity classification approach includes three classes for fatal and severe injury (KA) crashes, non-severe and possible injury (BC) crashes, and property damage only (PDO) crashes. While Granger Causality helped identify the most influential factors affecting crash severity, the learning-based models predicted the severity classes with varying performance. The results of Granger causality analysis identified the speed limit, surface and weather conditions, traffic volume, presence of workzones, workers in workzones, and high occupancy vehicle (HOV) lanes, among others, as the most important factors affecting crash severity. The prediction performance of the classifiers yielded varying results across the different classes. Specifically, while decision tree and random forest classifiers provided the greatest performance for PDO and BC severities, respectively, for the KA class, the rarest class in the data, deep neural net classifier performed superior than all other algorithms, most likely due to its capability of approximating nonlinear models. This study contributes to the limited body of knowledge pertaining to causal analysis and classification prediction of traffic crash injury severity using non-parametric approaches.

---

[1] * Corresponding author. Tel.: +1-480-634-3713;
*E-mail address:* chakra43@msu.edu

# 1. Introduction

The incredible economic and societal impact associated with traffic crashes provides motivation for transportation agencies to proactively pursue traffic safety improvements by reducing the frequency of traffic crashes and the degree of injury sustained by those involved in crashes. According to the national statistics presented annually by the Fatality Analysis Reporting System (FARS), in 2019, more than 36,000 people were killed and almost 2.74 million people sustained injuries in police-reported motor-vehicle crashes [1]. The background of this study is premised in Texas as it has historically been among the top states in terms of statewide fatalities in the U.S. As the data indicates, although Texas interstates account for only about 7% of the total interstate road-miles in the U.S. [2], the state alone accounts for greater than 10% of fatalities nationwide and only the interstates experience almost 17% of fatalities occurring across the state [3].

In the realm of traffic safety research, the development of reliable methodologies to predict and classify crash injury severity based on various explanatory variables has been a key component. Crash occurrence and injury result from a complex interaction among several factors including driver attributes, vehicle and traffic conditions, roadway geometric characteristics, features of the built environment, and weather conditions. A better understanding of the relationship between crashes (both frequency and injury severity) and these contributing factors can help identify appropriate countermeasures to improve roadway safety.

As found in the extant literature, previous research focusing on traffic safety analyses has widely utilized classical statistical techniques [4–7]. While these parametric methods have undoubtedly provided insights, the fundamental characteristics of crash data often result in methodological limitations that are not fully understood or accounted for. More recently, studies indicated that emerging data mining techniques offer superior prediction and greater accuracy due of their ability to work with massive amounts of multidimensional, outlying, and noisy data, modeling flexibility, and generalizability; outperforming the conventional statistical and econometric models [8, 9]. Also, the increasing availability of large-scale data from various sources, including connected and autonomous vehicles, and naturalistic driving studies, calls for a thorough understanding of the causal relationships between safety and various associated factors with the help of sophisticated methodological approaches.

## 1.1 Previous Work and Research Context

A significant body of research using regression models and related econometric techniques explored crash injury severity and an array of factors that influence it. Since injury severity data are generally represented by discrete categories (K or fatal, A or incapacitating injury, B or non-incapacitating injury, C or possible injury, and O or property damage only), largely discrete choice models, such as binary logit and probit models [10, 11], multinomial logit models [12, 13], and ordered response models [13, 14] have been applied to develop crash severity prediction models. Also, to account for unobserved heterogeneity in the data, mixed models are utilized [15]. As such, these parametric modeling techniques are subject to rather strict assumptions about the distribution of data and, usually, a linear functional form between dependent and explanatory variables. However, these assumptions may not always hold true. When basic assumptions of traditional statistical models were violated, erroneous estimations and incorrect inferences could be produced [16–18].

To overcome the limitations associated with traditional statistical models and the proficiency in capturing the nonlinear relationship between input and output data, more recently, researchers have proposed non-parametric methods and artificial intelligence models for crash injury severity analysis including decision trees (DT) [19], support vector machine (SVM) [9], and random forest (RF) [20], among others. Furthermore, as an emerging analytics technique, deep learning is increasingly being introduced into crash analysis [21]. These nonparametric analysis methods require no prior knowledge about crash factors and their functional relationships.

In an earlier study, Harb et al. (2009) employed the random forest modeling technique to interpret drivers', vehicles' and environmental characteristics associated with drivers' crash avoidance maneuvers. In another study, the prediction performance of random forest and decision trees were compared with a binary logit model for predicting drivers' gap decisions. The results illustrated that nonparametric data mining models were superior to the binary logit model at prediction accuracy [22]. Jeong et al. (2018) used five classification learning models including logistic regression, decision tree, neural network, gradient boosting model, and naïve Bayes classifier, to classify the injury severity in motor-vehicle crashes from across Michigan [23]. AlMamlook et al. (2019) compared different machine learning algorithms performance on predicting crash severity and found that the random forest algorithm outperformed logistic regression, naïve Bayes, and AdaBoost algorithms [20].

Chen et al. (2020) predicted severity of crashes from autonomous vehicles (AVs) in California using extreme gradient boosting (XGBoost) and classification and regression tree (CART). The study showed that the XGBoost model performed superior in identifying the injured crashes involving AVs. Also, the important features that contribute to the severity of crashes included weather, degree of vehicle damage, and crash location and type, among others [24]. Agarwal et al. (2019) utilized three non-parametric machine learning methods including classification and regression tree (CART), extreme gradient boosting (XGBoost), and support vector machine (SVM) to identify the critical factors affecting the severity of crashes on the state highways of India. Among the three studied methods, the XGBoost performed the best providing with high precision and fast processing speed as well as lower cost and complexity. The results show that the presence of pedestrian facility, type of collision, weather condition, intersection proximity, type of traffic control, and speed limits were critical factors affecting crash severity on state highways [25]. Zhang et al. (2018) determined that random forest model outperformed several other machine learning and statistical models including k-nearest neighbor, decision trees, support vector machine, ordered probit, and multinomial logit models in crash severity prediction [8].

In one of the recent studies using deep learning techniques, Zheng et al. (2019) investigated the use of convolutional neural networks (CNN) for crash severity prediction. When compared the prediction performance between the proposed CNN model and other machine learning models including k-nearest neighbor, decision trees, and support vector machine, results showed that CNN outperformed all other models [21]. Alkheder et al. (2017) used an artificial neural network (ANN) algorithm to predict traffic crash severity and compared its performance with an ordered probit (OP) model. Their results showed an improved accuracy achieved by the ANN model compared to that by the OP model [26]. The study by Arhin and Gatiba (2019) showed high accuracy in predicting injury severity with artificial neural network (ANN) models. However, they used only two severity levels, i.e. injury crash and non-injury crash, essentially combining fatal and severe injury crashes with non-serious and possible injury crashes into one class while predicting [27].

Crash severity datasets are typically imbalanced and highly skewed, where fatalities and severe injuries that are associated with significantly higher economic and societal costs, are much less represented in the data compared to other severities. In an imbalanced dataset, accuracy places more weight on the common classes than on the rarer classes. Therefore, the model's performance becomes worse for rarer classes. Hence, to improve the classification performance of the crash injury severity, it is crucial to develop models capable to deal with mislabels in the imbalanced crash data using robust analytical algorithms. Handling imbalanced data necessitates additional steps in model training and evaluation including balancing the dataset before training and using appropriate evaluation metrics. Yahayaa et al. (2020) confirmed that the mislabels in crash data significantly influence the injury severity predictions [28]. Jeong et al. (2018) used over- and under-sampling techniques to account for the imbalanced classes and determined that the classification performance was the highest when bagging is used with decision trees, with over-sampling treatment for imbalanced data [23]. Chen et al. (2020) applied synthetic minority oversampling technique (SMOTE) to deal with the imbalanced data [24].

In recent times, though there have been extensive research using big datasets, causal analysis for identifying causal structure and influential variables has received limited attention. Furthermore, there is no universally accepted definition of causality and the most commonly used notion is that of Granger causality, initially developed for analysis of econometric data [29]. Apart from Granger causality, Directed Information [30] and Transfer Entropy [31] are other definitions of causality that are used in literature. However, in dynamical systems setting all these measures fail to capture the correct causal structure [32] and, recently, a new notion of causality, known as information transfer, was defined to characterize and quantify causal structure and influence of a dynamical system [32–35]. Although there are different measures of causality, Granger causality is one of the most widely used methods, especially for static data, and it has been popular for identifying influential factors and prediction purposes in econometric studies and transportation research [36]. Beyzatlar et al. (2014) analyzed panel data from fifteen European countries (EU-15) using Granger causality to investigate the relationship between income and transportation [37]. In order to accurately build the traffic flow prediction models, Li et al. (2015) utilized Granger causality to determine the potential dependence among the pool of predictor variables in the time-series big data collected by different sensors [38]. Ageli and Zaidan (2013) used Granger causality to examine the association between traffic crashes and other relevant factors [39].

To summarize, significant attention has been given to crash severity modeling in previous research, especially with parametric methods, but injury outcome prediction and evaluating the causal structure of traffic crashes were not studied adequately. To this end, this study presents a methodological framework to model the severity of motor vehicle crashes on interstates. The analysis involves causal inference, using Granger causality tests and injury severity classification using different machine learning and deep learning approaches including decision trees (DT), random forest (RF), extreme gradient boosting (XGBoost), and deep neural network (DNN). While Granger Causality helped identify the important factors affecting crash severity, the learning-based models predicted the severity classes with varying performance. The methodological framework in this study is designed to incorporate a broad range of potential factors influencing crash severity including driver-, environment-, traffic-, and roadway-related characteristics. The output of the proposed crash severity classification approach includes three classes: fatal and severe injury (KA) crashes, non-severe and possible injury (BC) crashes, and property damage only (PDO) crashes.

## 2. Methodology

### 2.1 Data

The collection and compilation of data for this analysis was an evolving process. The data on traffic crashes were obtained from the crash database maintained by Texas Department of Transportation (TxDOT), known as Crash Records Information System (CRIS) [40]. In CRIS, as is standard, each crash was assigned to the most severe level experienced by any vehicle involved. In addition to the crash severity level, the database provides several other relevant information including but not limited to roadway class and system, crash location and type, crash time and year, manner of collision, statutory and suggested speed limits, traffic volumes, roadway cross-sectional characteristics, and surface, lighting, and weather conditions. The traffic volumes in the crash database were from 2018. Therefore, the growth factors were applied from the Traffic Count Database System (TCDS) maintained by TxDOT to obtain traffic volumes of other years [41].

For this research, crash data from all interstates across the state were collected for a period of six years, between 2014 and 2019. Prior research indicated that underlying factors influencing traffic safety largely vary between land use types (rural, urban, and suburban) and roadway functional classes [42, 43]. Hence, for the purpose of this study, crashes occurring only on interstates in mostly urban areas (population more than 50,000) and a few suburban areas on the fringes of major cities with population between 25,000 and 49,000 were evaluated. This resulted in a total of 156,166 crashes occurring across 59 counties out of a total of 254 counties statewide during the analysis period. Observably, in all years, the proportions of crashes increase as the severity of crashes decreases. Overall, property damage only (PDO) crashes were the most frequent type of crashes.

### 2.2 Causal Analysis

Granger causality test is a statistical hypothesis test which determines whether one time series is helpful in predicting another time series. In particular, a variable *X* Granger causes a variable *Y* if the prediction of *Y*, based on its own past and past of *X* is different than the prediction of *Y* based on its own past alone. For the self-containment of the paper, the basic formulation of Granger causality is briefly discussed.

The notion of Granger causality assumes that cause happens before the effect and the cause has unique information about the future of the effect. With this, Granger causality can be formally defined as,
**Definition:** The hypothesis for Granger causality of *X* on *Y* is
$$P[Y(t+1) \in \Omega | I(t)] \neq P[Y(t+1) \in \Omega | I_{\bar{x}}(t)]$$
where $P[Y(t+1) \in \Omega | I(t)]$ is the probability of *Y(t+1)* belonging to the set Ω when the entire information till time *t* is considered (I(t)) and $P[Y(t+1) \in \Omega | I_{\bar{x}}(t)]$ is the probability of *Y(t+1)* belonging to the set Ω when *X* is removed from the information set (denoted by $I_{\bar{x}}(t)$). When above hypothesis is satisfied, it is said *X* Granger causes *Y*.

Though the above definition is for a bivariate time-series, Granger causality test can be used for causal discovery among variables of a multivariate time-series data as well. The naïve way to do that is to consider two variables at a time, but this often leads to erroneous conclusions [44, 45] and for causal discovery from a multivariate time-series data, the relevant idea is the concept of conditional Granger causality.

Suppose *X*, *Y* and *Z* are three jointly distributed multivariate stochastic processes and consider the regression models

$$X_t = \alpha_t + \left(X_{t-1}^{(p)} \oplus Z_{t-1}^{(r)}\right) \cdot A + \epsilon_t \qquad (1)$$

$$X_t = \alpha_t' + \left(X_{t-1}^{(p)} \oplus Y_{t-1}^{(q)} \oplus Z_{t-1}^{(r)}\right) \cdot A' + \epsilon_t' \qquad (2)$$

where *A* and *A'* are the regression coefficients, $\alpha$ and $\alpha'$ are constants and $\epsilon_t$ and $\epsilon_t'$ are the residuals. The predictee variable *X* is first regressed on previous *p* lags of itself and *r* lags of the conditional variable *Z* and second on previous *p* lags of itself, *q* lags of *Y* and *r* lags of *Z*. With this, the Granger causality of *Y* on *X*, given *Z*, is

$$G_{Y \to X | Z} = \ln \frac{var(\epsilon_t)}{var(\epsilon'_t)}$$

where $var(\cdot)$ denotes the variance, and $G_{Y \to X|Z}$ is a measure of the extent to which the inclusion of *Y* in the model (2) reduces the prediction error of (1).

### 2.3 Machine Learning Algorithms for Classification

In this section, the different machine learning techniques that have been used in this study are briefly discussed. Note that, in this study, a multi-class classification problem has been addressed using four different methods, namely, Decision Trees, Random Forests, XGBoost Classifier and finally Deep Neural Nets.

*Decision Trees:* Decision tree is a supervised learning algorithm which has a tree structure where at each node a yes-no kind of question is answered (Figure 1). Hence, unlike probabilistic classification algorithms, decision tree is a decision or rule-based algorithm. For decision tree classification, one uses the dataset features to create yes-no questions and this process continuously splits the dataset until all the data-points belonging to each class are isolated. The process of asking one yes-no question is equivalent to adding one node to the decision tree and the result of asking the question splits the dataset according to the feature. As shown in Figure 1, the root node is the first node of the tree, the leaf nodes correspond to nodes where one decides to stop after a split and the in-between nodes are known as decision nodes. The decision tree algorithm tries to completely divide the dataset such that to each leaf node the algorithm assigns the most common class among all the data points in that node.

*Random Forest:* Random Forest is an ensemble classification algorithm which builds a collection of decision trees and is usually trained using a bagging algorithm. As in Figure 2, multiple decision trees are trained separately and the output of the random forest is obtained as an average of the outputs of individual decision trees.

*XGBoost Classifier:* XGBoost, an ensemble algorithm, stands for extreme boosting in the sense that it *boosts* the performance of a regular gradient boosting algorithm. In particular, XGBoost creates a sequence of models that sequentially corrects the model obtained from the previous step. That is, the first model that is built on the training data is corrected (updated) by the second model, which in turn in improved upon by the third model and so on and so forth. The main advantage XGBoost classifier over other ensemble classifiers is the fact that it is comparatively faster than other methods. Furthermore, the core algorithm can be parallelized and hence it can take advantage of multi-core processors. It can also be parallelized over GPUs and network of computers, thus making it feasible for really large datasets.

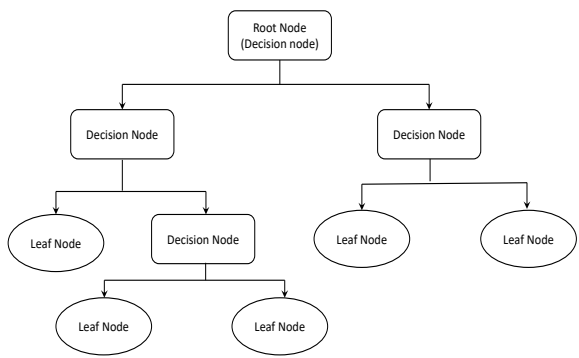

**Figure 1: Decision Tree flow diagram.**

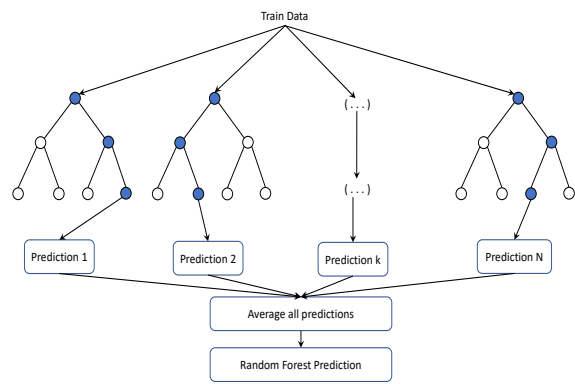

**Figure 2: Random Forest flow diagram.**

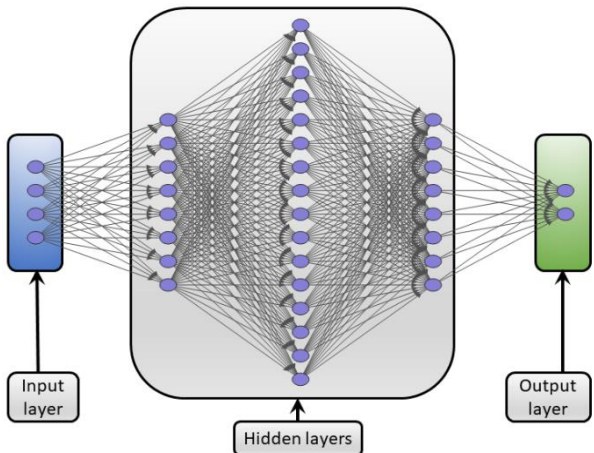

*Deep Neural Network Classifier:* Deep Neural Network is an artificial neural network with multiple *hidden* layers between the inputs and outputs. For example, in Figure 3, the input layer has 4 nodes (neurons) and the output layer has 2 nodes. The three in-between layers, with 8, 16 and 8 nodes (neurons) are the hidden layers. For classification problems, the number of nodes in the output layer correspond to the number of classes. One of the main advantages of a Neural Net is the fact that it can approximate a nonlinear function to arbitrary accuracy and this can be achieved by properly choosing the activation functions. Although the choice of activation function depends on the task, Rectified Linear Unit, Hyperbolic Tangent and Sigmoid are among the mostly used activation functions.

**Figure 3: A Deep Neural Network (DNN) with three hidden layers.**

## 3. Analysis and Results

### 3.1 Data Summary and Preliminary Analysis

The summary statistics of the data prior to addressing the data imbalance issue are summarized in Table 1. The minimum and maximum are provided for each variable, along with their respective mean, and standard deviation. As can be seen from Table 1, the traffic volumes vastly vary across the freeways. The crash locations also include work zones, hence for approximately 2% of crashes, the speed limit was 45 mph, the minimum for this factor. Also, almost 93% of locations were in urban areas with population greater than 50,000, while the remaining crash locations were from suburban areas. Consistent with urban nature of roadways, more than a quarter of observations have HOV lanes, and in most cases, more than 4 lanes, considering both traffic directions.

**Table 1: Descriptive Statistics of the Data**

| Parameter | Min. | Max. | Mean | S.D. |
|---|---|---|---|---|
| Fatal and Severe Injury Crashes (KA) | 0 | 1 | 0.03 | 0.16 |
| Non-severe and Possible Injury Crashes (BC) | 0 | 1 | 028 | 0.45 |
| Property Damage Only Crashes (O) | 0 | 1 | 0.69 | 0.46 |
| Annual Average Daily Traffic (AADT) (vehicles per day) | 4,606 | 330,096 | 144,961 | 63,631 |
| Speed Limit (mph) | 45 | 80 | 63.09 | 5.96 |
| Proportion of Heavy Vehicle (%) | 3.7 | 95.08 | 11.49 | 6.80 |
| Single Vehicle Crashes (SV) | 0 | 1 | 0.16 | 0.37 |
| Work Zone Presence | 0 | 1 | 0.13 | 0.34 |
| Worker Prersent in the Work Zone | 0 | 1 | 0.05 | 0.21 |
| Vurnerable Road User Involved | 0 | 1 | 0.004 | 0.06 |
| Number of Lanes > 4 | 0 | 1 | 0.84 | 0.36 |
| Commercial Vehicle Involved | 0 | 1 | 0.15 | 0.36 |
| Presence of High Occupancy Vehicle Lanes | 0 | 1 | 0.27 | 0.44 |
| Population > 50,000 | 0 | 1 | 0.93 | 0.16 |
| Crash Time = Peak Hour | 0 | 1 | 0.29 | 0.46 |
| Dry Surface | 0 | 1 | 0.83 | 0.37 |
| Clear Weather | 0 | 1 | 0.71 | 0.45 |
| Day light or dark Unlighted | 0 | 1 | 0.90 | 0.30 |
| Curved Road Alignment | 0 | 1 | 0.15 | 0.23 |
| Left Turning Curved Road | 0 | 1 | 0.10 | 0.30 |
| Spiral Curved Road | 0 | 1 | 0.01 | 0.08 |
| Median Barrier Not Present | 0 | 1 | 0.01 | 0.12 |
| Median Width < 12 feet | 0 | 1 | 0.44 | 0.50 |
| No Left Shoulder Present | 0 | 1 | 0.09 | 0.29 |
| Left Shoulder Width < 6 feet | 0 | 1 | 0.03 | 0.16 |
| No Right Shoulder Present | 0 | 1 | 0.01 | 0.03 |
| Right Shoulder Width < 6 feet | 0 | 1 | 0.001 | 0.01 |

### 3.2 Granger Causality

A total of 24 variables were evaluated using Granger causality and a set of most influencing predictors were identified and ranked, based on the causality scores (Figure 4). The optimal lag for the VAR model for Granger causality analysis was evaluated using AIC criterion and was determined to be 4. Ultimately, a total of 17 predictors with the highest scores were selected for the classification analysis and compared that with the full set of variables. For the most variables, the rank of importance makes sense and compare favorably with the earlier research [46, 47], indicating speed limits, traffic volumes, percent of heavy vehicles, or presence of workzones to be some of the most important factors in classifying crash injury severity. The Granger causality analysis was performed on R statistical software version 4.0.2.

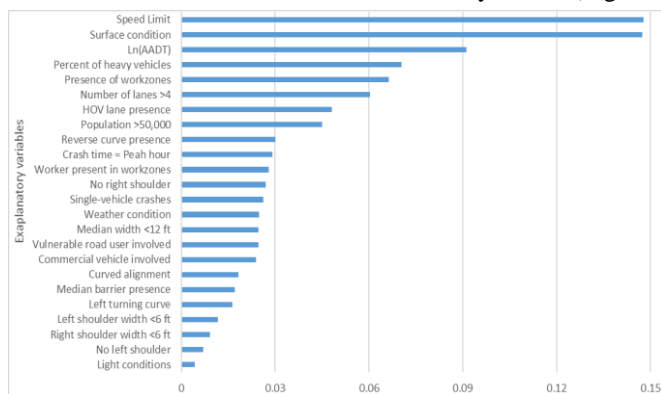

**Figure 4: Rank ordering of independent variables according to Granger causality test**

### 4. Classification Results and Discussion

The classification of crash injury severity utilizing different machine learning algorithms is carried out with the 17 most important predictors as selected based on the causality score and its prediction performance is compared with that using all variables. The details of the classification analysis and results are discussed in the following sub-sections. The classification analysis was carried out in Python 3.8 and Tensorflow 2.7.

*Data Balancing and Pre-processing:* The classification of crash injury severity in this study considered three different injury classes including fatal and severe injury (KA), non-severe and possible injury (BC), and no injury or property damage only (O or PDO) from the traditional KABCO scale. For training and testing purposes, the entire dataset was split into two parts, where 80% of the data was used for training the models, while the remaining 20% data was used for testing. As is commonly the case with most crash data, the data analyzed in this study was highly skewed, in which the number of observations for no injury class is more than 30 times the number of observations in fatal and severe injury class. To address the issue with biased predictions due to imbalanced data, the training data was balanced using random under-sampling of no injury class observations and over-sampling of the data of the remaining classes. This was done by using the Synthetic Minority Oversampling Technique (SMOTE) technique in python.

*Classification using Machine Learning Algorithms:* This study utilized four different machine learning algorithms, namely, Decision Trees, Random Forest, Extreme Gradient Boosting, and Deep Neural Network for classifying the crash injury severity. Usually, the performance of a classifier is quantified in terms of the accuracy, f1 score, or AUC score. However, since the data analyzed here has a high degree of imbalance, it may happen that for a classifier model these scores are high, but it fails to predict the minority classes. Hence, the confusion matrix for each classifier is considered as a measure of its efficiency. In all cases, the normalized confusion matrices are presented for the models using the 17 most influential variables selected through Granger causality (reduced classifier) and are compared with those using all the independent variables (full classifier).

The normalized confusion matrices of the Decision Tree (DT) classifier in Figure 5 show that the true positives for both KA (Label 2) and BC (Label 1) classes decrease with reduced classifier, while that for PDO (Label 0) class increases. Additionally, the computational complexity and training time are lower when fewer number of features are considered compared to all predictors. These results also confirm that causal analysis of a dataset not only identifies the most influential variables appropriately, thus providing a deeper insight into the data and the problem at hand, it can also be used to select a subset of independent variables while developing a model (classification or regression) which reduces the computational complexity, without compromising much on the predictive performance.

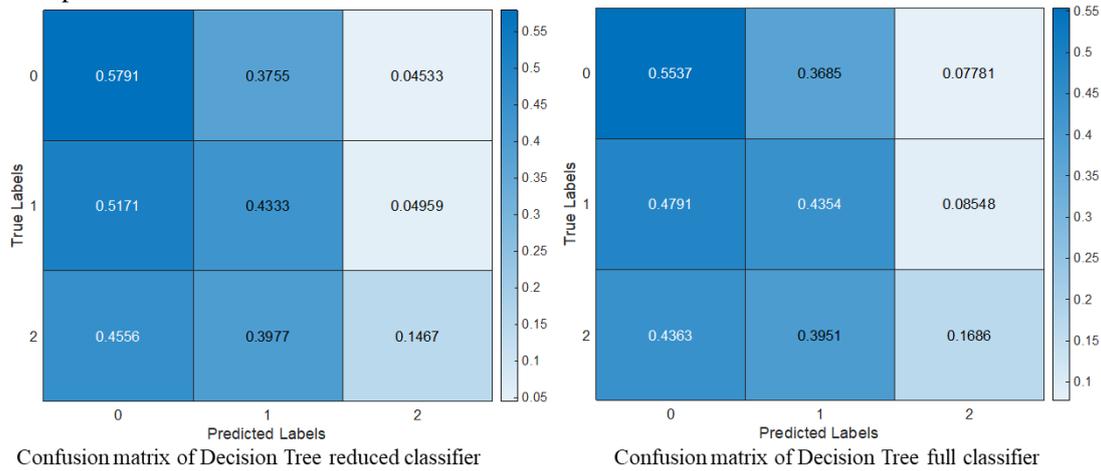

Confusion matrix of Decision Tree reduced classifier　　　Confusion matrix of Decision Tree full classifier

**Figure 5: Confusion matrix of Decision Tree classifier**

Similar observations can be made from the normalized confusion matrices of the Random Forest (RF) classifier (Figure 6) where we trained the data with the number of estimators set to 1,000. As with the DT classifier, for the RF classifier the classification performance for the PDO crashes improves for the reduced classifier, whereas for the BC class, it remains the same and for KA class, the performance of the reduced classifier degrades.

Next, the normalized confusion matrices for the Extreme Gradient Boosting (XGBoost) classifier in Figure 7 shows a considerable improvement in prediction performance for KA class compared to DT and RF classifiers. Moreover, for the XGBoost classifier, the predictive performance of the reduced classifier improves substantially for PDO class, while for BC class, it remains almost the same for both XGBoost models.

The final classifier considered in this study was a Deep Neural Network (DNN) with all-to-all connections between the consecutive layers. Four hidden layers were considered with 128 nodes for the first three hidden layers and 64 nodes for the layer. The activation functions used were Rectified Linear Units (ReLU) for the hidden layers and softmax for the output layer. With this, the DNN was trained for 150 epochs with a batch size of 2,048. The confusion matrices for the trained models are presented in Figure 8.

It is important to note that the performance of different classifiers varied across the different injury severity classes, a finding consistent with previous studies [48]. When considering DT, RF, and XGBoost, for PDO class, DT performs the best on the test set among the three different classifiers, followed by similar performances of RF and XGBoost classifiers. For BC class, RF performs the best, followed by XGBoost and then DT classifiers. The fatal and severe injury (KA) crashes are the costliest and of greatest importance to transportation safety researchers and engineers. Thus, the classifier that provides the maximum true positives for KA class should be chosen. Among DT, RF, and XGBoost classifiers, the highest true positives for KA class is provided by the XGBoost classifier outperforming both DT and RF classifiers. However, the prediction performance for KA class degrades by a significant margin for the reduced XGBoost classifier. This is because, while identifying and rank ordering the independent variables according to the Granger causality score, the entire dataset was considered which was highly imbalanced. This makes any model, trained on this imbalanced data, highly biased to the larger class and as such the ordering of the influential variables mostly capture the influential variables for PDO class. This is reflected in all the classifier models where the predictive performance for PDO class increases when only the most influential variables are considered. This strongly affirms that the factors which *causes* more severe crashes are different that those causing lower or no injury crashes. Finally, based on the number of true positives, although the DNN classifier compares favorably with DT, RF, and XGBoost classifiers for BC and PDO classes, it outperforms all other classifiers as far as correct prediction for KA class is concerned. This is most likely because, depending on the architecture of DNN, it can learn highly nonlinear models.

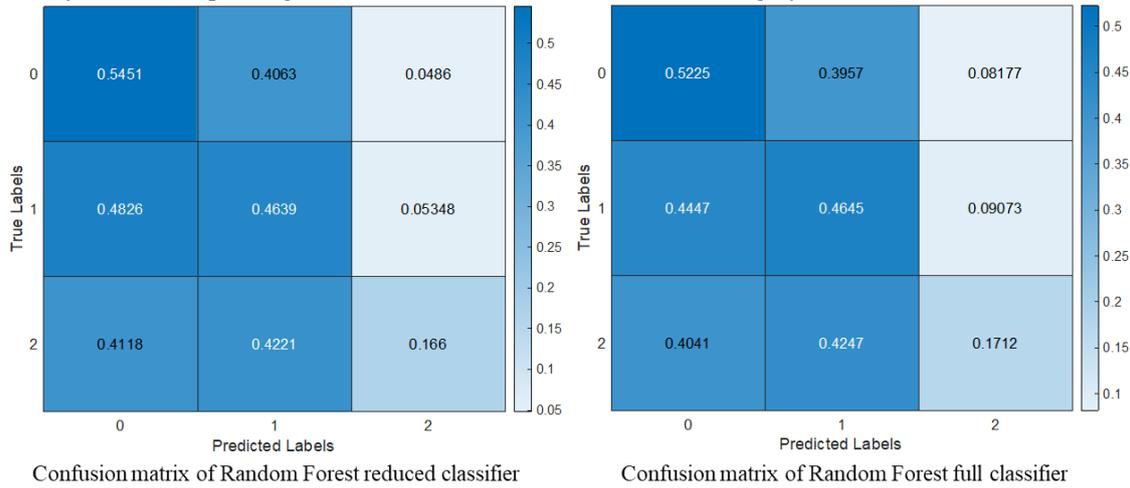

Confusion matrix of Random Forest reduced classifier     Confusion matrix of Random Forest full classifier
**Figure 6: Confusion matrix of Random Forest classifier**

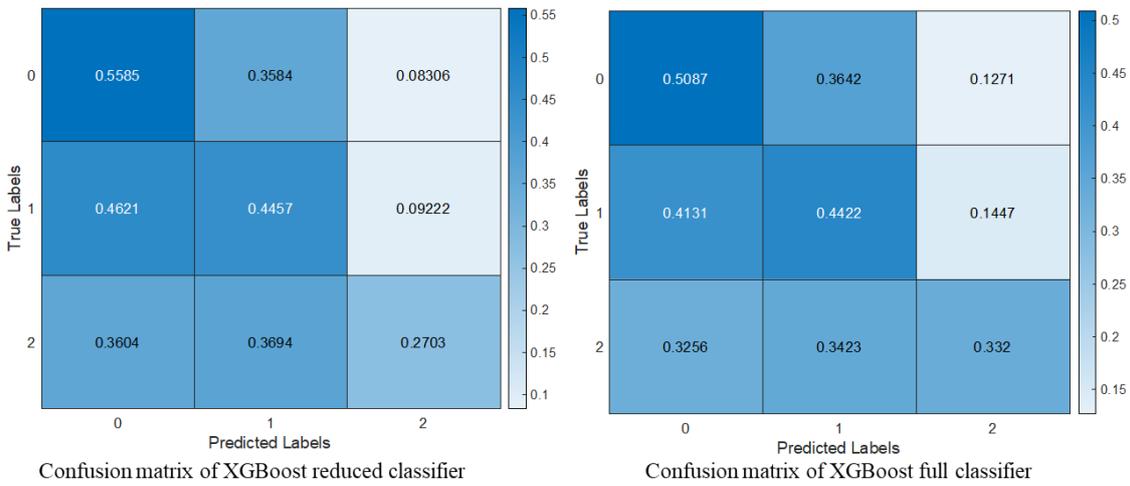

Confusion matrix of XGBoost reduced classifier     Confusion matrix of XGBoost full classifier
**Figure 7: Confusion matrix of XGBoost classifier**

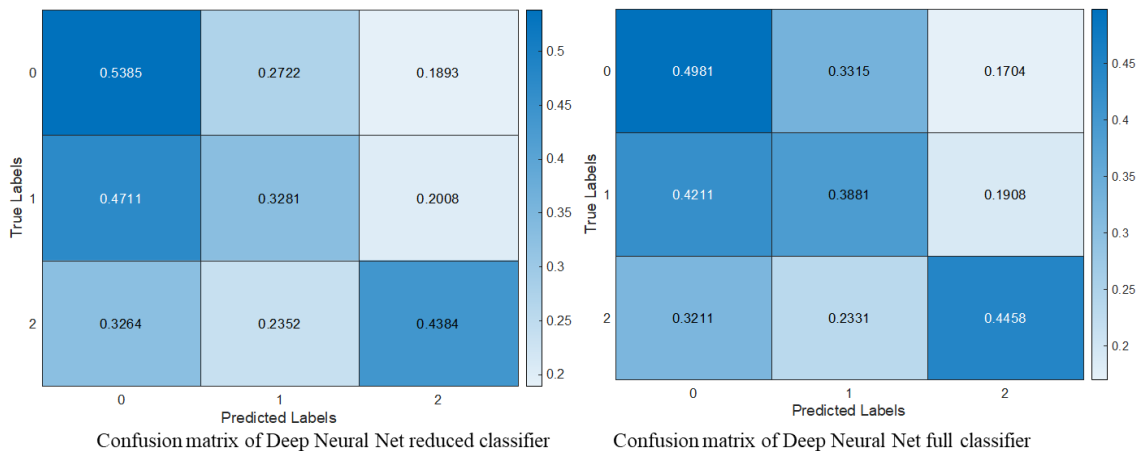

Confusion matrix of Deep Neural Net reduced classifier     Confusion matrix of Deep Neural Net full classifier
**Figure 8: Confusion matrix of Deep Neural Net classifier**

## 5. Conclusions

This study presents a methodological framework involving the development of causal inference and injury severity classification for freeway traffic crashes. Granger causality test was used to identify and rank-order the influential features causing the varying injury severity that were further utilized to build four different classifiers, including Decision Tree (DT), Random Forest (RF), Extreme Gradient Boosting (XGBoost), and Deep Neural Network (DNN) classifiers to classify the crashes according to their severity. The output of the proposed classification approach includes three severity classes for fatal and severe injury (KA) crashes, non-severe and possible injury (BC) crashes, and property damage only (PDO) crashes.

The most influencing factors identified by Granger causality include speed limit, surface and weather conditions, traffic volume, presence of workzones, workers in workzones, and HOV lanes, among others. The causal analysis not only provides greater insight into the dataset and the problem at hand, but also provides a basis for choosing the *right* variables for constructing reduced order models. Granger causality test provides a systematic procedure for selecting the influential variables and as shown in this study, the efficacy of the Granger causality was demonstrated by achieving comparable results between reduced order and full order models. In terms of the prediction performance of the classifiers, decision tree and random forest classifiers provided the greatest performance for PDO and BC crash severities, respectively. For the KA class, the rarest class in the data, deep neural net classifier performed superior to all other algorithms, most likely due to its capability in approximating nonlinear models.

The study identifies some limitations that can be addressed in future research. Firstly, the factors affecting injury severity often differ based on severity level and this was not considered in this study wherein the influential factors were identified using all severity classes together. In future, identification of important factors separately for different injury levels would be more insightful. Additionally, to improve the prediction performance further, a subsequent analysis can be carried out by using binary classifiers for fatal and injury, and PDO crashes. Nevertheless, overall, this study contributes to the limited body of knowledge pertaining to causal analysis and classification prediction of traffic crash injury severity using non-parametric approaches.

## References

1 Crashes by Crash Severity, NHTSA, https://cdan.nhtsa.gov/SASStoredProcess/guest, accessed October 2021
2 Public Road Length 2019. Miles by Functional System. Table HM 20., https://www.fhwa.dot.gov/policyinformation/statistics/2019/pdf/hm20.pdf, accessed November 2021
3 Texas Fatal Crashes and Fatalities by Month and Road Type 2019, https://ftp.txdot.gov/pub/txdot-info/trf/crash_statistics/2019/05.pdf
4 Chakraborty, M., Gates, T.J. Association between Driveway Land Use and Safety Performance on Rural Highways. Transportation Research Record, 2020
5 Chakraborty, M., Stapleton, S.Y., Ghamami, M., Gates, T.J. Safety Effectiveness of All-Electronic Toll Collection systems. Advances in Transportation Studies, 2020, 2, (Special Issue), pp. 127–142
6 Chakraborty, M., Singh, H., Savolainen, P.T., Gates, T.J. Examining Correlation and Trends in Seatbelt Use among Occupants of the Same Vehicle using a Bivariate Probit Model. Transportation Research Record, 2021
7 Chakraborty, M., Mahmud, S., Gates, T. Analysis of Trends and Correlation in Child Restraint Use and Seating Position of Child Passengers in Motor Vehicles: Application of a Bivariate Probit Model. Transportation Research Record: Journal of the Transportation Research Board, 2022
8 Zhang, J., Li, Z., Pu, Z., Xu, C. Comparing Prediction Performance for Crash Injury Severity Among Various Machine Learning and Statistical Methods. IEEE Access, 2018, 6, pp. 60079–60087
9 Iranitalab, A., Khattak, A. Comparison of four statistical and machine learning methods for crash severity prediction. Accident Analysis & Prevention, 2017, 108, pp. 27–36
10 Haleem, K., Abdel-Aty, M. Examining traffic crash injury severity at unsignalized intersections. Journal of Safety Research, 2010, 41, (4), pp. 347–357
11 Kononen, D.W., Flannagan, C.A.C., Wang, S.C. Identification and validation of a logistic regression model for predicting serious injuries associated with motor vehicle crashes. Accident Analysis & Prevention, 2011, 43, (1), pp. 112–122
12 Malyshkina, N.V., Mannering, F.L. Empirical assessment of the impact of highway design exceptions on the frequency and severity of vehicle accidents. Accident Analysis & Prevention, 2010, 42, (1), pp. 131–139
13 Ye, F., Lord, D. Comparing three commonly used crash severity models on sample size requirements: Multinomial logit, ordered probit and mixed logit models. Analytic Methods in Accident Research, 2014, 1, pp. 72–85
14 Xie, Y., Zhang, Y., Liang, F. Crash Injury Severity Analysis Using Bayesian Ordered Probit Models. J. Transp. Eng., 2009, 135, (1), pp. 18–25
15 Mannering, F.L., Shankar, V., Bhat, C.R. Unobserved heterogeneity and the statistical analysis of highway accident data. Analytic Methods in Accident Research, 2016, 11, pp. 1–16
16 Mussone, L., Ferrari, A., Oneta, M. An analysis of urban collisions using an artificial intelligence model. Accident Analysis & Prevention, 1999, 31, (6), pp. 705–718
17 Delen, D., Sharda, R., Bessonov, M. Identifying significant predictors of injury severity in traffic accidents using a series of artificial neural networks. Accident Analysis & Prevention, 2006, 38, (3), pp. 434–444

18. Chakraborty, M., Mahmud, S., Gates, T., Sinha, S. Linear Regularization-based Analysis and Prediction of Human Mobility in the U.S. during the COVID-19 Pandemic. engrXiv, 2020
19. de Oña, J., López, G., Abellán, J. Extracting decision rules from police accident reports through decision trees. Accident Analysis & Prevention, 2013, 50, pp. 1151–1160
20. AlMamlook, R.E., Kwayu, K.M., Alkasisbeh, M.R., Frefer, A.A. Comparison of Machine Learning Algorithms for Predicting Traffic Accident Severity, in 2019 IEEE Jordan International Joint Conference on Electrical Engineering and Information Technology (JEEIT), IEEE, 2019, pp. 272–276
21. Zheng, M., Li, T., Zhu, R., Chen, J., Ma, Z., Tang, M., Cui, Z., Wang, Z. Traffic Accident's Severity Prediction: A Deep-Learning Approach-Based CNN Network. IEEE Access, 2019, 7, pp. 39897–39910
22. Harb, R., Yan, X., Radwan, E., Su, X. Exploring precrash maneuvers using classification trees and random forests. Accident Analysis & Prevention, 2009, 41, (1), pp. 98–107
23. Jeong, H., Jang, Y., Bowman, P.J., Masoud, N. Classification of motor vehicle crash injury severity: A hybrid approach for imbalanced data. Accident Analysis & Prevention, 2018, 120, pp. 250–261
24. Chen, H., Chen, H., Liu, Z., Sun, X., Zhou, R. Analysis of Factors Affecting the Severity of Automated Vehicle Crashes Using XGBoost Model Combining POI Data. Journal of Advanced Transportation, 2020
25. Agarwal, V., Chatterjee, S., Mitra, S. Crash severity analysis through nonpaametric machine learning methods. Journal of Eastern Asia Society for Transportation Studies, 2019, 13
26. Alkheder, S., Taamneh, M., Taamneh, S. Severity Prediction of Traffic Accident Using an Artificial Neural Network: Traffic Accident Severity Prediction Using Artificial Neural Network. J. Forecast., 2017, 36, (1), pp. 100–108
27. Arhin, S.A., Gatiba, A. Predicting Injury Severity of Angle Crashes Involving Two Vehicles at Unsignalized Intersections Using Artificial Neural Networks. Eng. Technol. Appl. Sci. Res., 2019, 9, (2), pp. 3871–3880
28. Yahaya, M., Fan, W., Fu, C., Li, X., Su, Y., Jiang, X. A machine-learning method for improving crash injury severity analysis: a case study of work zone crashes in Cairo, Egypt. International Journal of Injury Control and Safety Promotion, 2020, 27, (3), pp. 266–275
29. Granger, C.W.J. Investigating Causal Relations by Econometric Models and Cross-spectral Methods. Econometrica, 1969, 37, (3), pp. 424–438
30. Massey, J.L. Causality, Feedback, and Directed Information. International Symposium on Information Theory and its Applications, 1990
31. Schreiber, T. Measuring Information Transfer. Phys. Rev. Lett., 2000, 85, (2), pp. 461–464
32. Sinha, S., Vaidya, U. Causality preserving information transfer measure for control dynamical system, in '2016 IEEE 55th Conference on Decision and Control (CDC)' 2016 IEEE 55th Conference on Decision and Control (CDC), IEEE, 2016, pp. 7329–7334
33. Sinha, S., Vaidya, U. On information transfer in discrete dynamical systems, in '2017 Indian Control Conference (ICC)' 2017 Indian Control Conference (ICC), IEEE, 2017, pp. 303–308
34. Sinha, S., Sharma, P., Vaidya, U., Ajjarapu, V. Identifying causal interaction in power system: Information-based approach, in '2017 IEEE 56th Annual Conference on Decision and Control (CDC)' 2017 IEEE 56th Annual Conference on Decision and Control (CDC), IEEE, 2017, pp. 2041–2046
35. Sinha, S., Sharma, P., Vaidya, U., Ajjarapu, V. On Information Transfer-Based Characterization of Power System Stability. IEEE Trans. Power Syst., 2019, 34, (5), pp. 3804–3812
36. Sinha, S., Chakraborty, M. Causal Analysis and Prediction of Human Mobility in the U.S. during the COVID-19 Pandemic. arXiv:2111.12272 [cs, stat], 2021
37. Beyzatlar, M.A., Karacal, M., Yetkiner, H. Granger-causality between transportation and GDP: A panel data approach. Transportation Research Part A: Policy and Practice, 2014, 63, pp. 43–55
38. Li, L., Su, X., Wang, Y., Lin, Y., Li, Z., Li, Y. Robust causal dependence mining in big data network and its application to traffic flow predictions. Transportation Research Part C: Emerging Technologies, 2015, 58, pp. 292–307
39. Ageli, M.M., Zaidan, A.M. Road Traffic Accidents in Saudi Arabia: An ADRL Approach and Multivariate Granger Causality. IJEF, 2013, 5, (7)
40. CRIS Query, https://cris.dot.state.tx.us/public/Query/app/public/welcome, accessed April 2019
41. Traffic Count Database System (TCDS), https://txdot.ms2soft.com/tcds/tsearch.asp?loc=Txdot&mod=TCDS, accessed April 2019
42. Chakraborty, M., Gates, T. Assessing Safety Performance on Urban and Suburban Roadways of Lower Functional Classification: A Comparison of Minor Arterial and Collector Roadway Segments. Transportation Research Record: Journal of the Transportation Research Board, 2022
43. Chakraborty, M., Gates, T.J. Relationship between Horizontal Curve Density and Safety Performance on Rural Two-Lane Road Segments by Road Jurisdiction and Surface Type, in TRB Annual Meeting, 2021
44. Ding, M., Chen, Y., Bressler, S.L. Granger Causality: Basic Theory and Application to Neuroscience. arXiv:q-bio/0608035, 2006
45. Geweke, J.F. Measures of Conditional Linear Dependence and Feedback between Time Series. Journal of the American Statistical Association, 1984, 79, (388), pp. 907–915
46. Elvik, R., Vaa, T., Hoye, A., Sorensen, M. The Handbook of Road Safety Measures. Emerald Group Publishing, 2009
47. Osman, M., Paleti, R., Mishra, S. Analysis of passenger-car crash injury severity in different work zone configurations. Accident Analysis & Prevention, 2018, 111, pp. 161–172
48. Ahmadi, A., Jahangiri, A., Berardi, V., Machiani, S.G. Crash severity analysis of rear-end crashes in California using statistical and machine learning classification methods. Journal of Transportation Safety & Security, 2020, 12, (4), pp. 522–546